# MIR: Efficient Exploration in Episodic Multi-Agent Reinforcement Learning via Mutual Intrinsic Reward

Kesheng Chen[1], Wenjian Luo[1,2], Bang Zhang[1], Zeping Yin[1] and Zipeng Ye[1]

[1] Guangdong Provincial Key Laboratory of Novel Security Intelligence Technologies, School of Computer Science and Technology, Harbin Institute of Technology, Shenzhen 518055, China

[2] Peng Cheng Laboratory, Shenzhen 518005, China.

`22s151138@stu.hit.edu.cn, luowenjian@hit.edu.cn,`

**Abstract.** Episodic rewards present a significant challenge in reinforcement learning. While intrinsic reward methods have demonstrated effectiveness in single-agent reinforcement learning scenarios, their application to multi-agent reinforcement learning (MARL) remains problematic. The primary difficulties stem from two factors: (1) the exponential sparsity of joint action trajectories that lead to rewards as the exploration space expands, and (2) existing methods often fail to account for joint actions that can influence team states. To address these challenges, this paper introduces Mutual Intrinsic Reward (MIR), a simple yet effective enhancement strategy for MARL with extremely sparse rewards like episodic rewards. MIR incentivizes individual agents to explore actions that affect their teammates, and when combined with original strategies, effectively stimulates team exploration and improves algorithm performance. For comprehensive experimental validation, we extend the representative single-agent MiniGrid environment to create MiniGrid-MA, a series of MARL environments with sparse rewards. Our evaluation compares the proposed method against state-of-the-art approaches in the MiniGrid-MA setting, with experimental results demonstrating superior performance.

**Keywords:** Multi Agent Reinforce Learning, Intrinsic Reward

## 1    Introduction

Sparse-reward reinforcement learning extensively employs auxiliary models trained on interaction data (e.g., agent observations/actions) to generate intrinsic rewards for exploration. These models quantify environmental novelty via prediction error (ICM [16], NGU [2]) or state-count metrics [15]. Advanced approaches integrate feature extractors (CNNs/RNNs) to enhance environmental change detection, with methods like RND [4], NovelD [29], and DEIR [24] evaluating state novelty through embedding comparisons. DEIR notably achieves state-of-the-art results on MiniGrid [5] tasks.

However, applying these methods to sparse-reward multi-agent RL (MARL) faces critical challenges: (1) exponential growth of state/action spaces with agent numbers, and (2) shared novelty metrics impeding individualized reward allocation. Current

MARL strategies prioritize single-agent behavior optimization, failing to coordinate multi-agent optimal actions.

This work addresses these limitations through the Centralized Training with Decentralized Execution (CTDE) paradigm (e.g., MAPPO [28], VDN [22]), which enhances global information utilization. We propose a Mutual Intrinsic Reward (MIR) mechanism that quantifies inter-agent observation changes via auxiliary models. By assigning individual rewards based on teammates' observational novelty, MIR synergizes with existing exploration methods, significantly improving team performance in sparse-reward MARL scenarios.

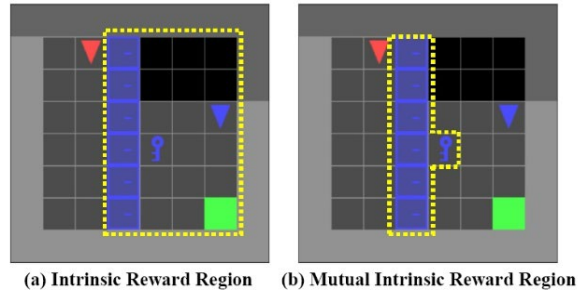

(a) Intrinsic Reward Region     (b) Mutual Intrinsic Reward Region

**Fig. 1.** Illustration of different reward focus areas

Figure 1 depicts a two-agent maze with sparse rewards, where the blue agent must unlock a door (blue) to enable the red agent's access to the endpoint (red rectangle). Global rewards occur only when all agents reach the endpoint. Unlike conventional methods (e.g., DEIR) targeting unexplored areas (yellow dashed), MIR prioritizes regions inducing teammate observation changes (e.g., door unlocking alters the red agent's state). This motivates our flexible mutual reward framework, enhancing team exploration efficiency in sparse-reward MARL.

We extend the MiniGrid [5] benchmark with MiniGrid-MA, a specialized MARL suite. Evaluations compare our mutual methods (DEIR-MIR, NovelD-MIR) with state-of-the-art (SOTA) intrinsic reward approaches. Results confirm significant performance gains in cooperative sparse-reward tasks.

In summary, this paper makes three key contributions:

- Implementation of existing intrinsic reward methods within CTDE-compliant MARL algorithms, coupled with the creation of the MiniGrid-MA environment specifically designed for sparse-reward multi-agent evaluation.
- Proposal of a novel mutual intrinsic reward (MIR) framework that effectively enhances existing intrinsic reward methods in MARL settings. This includes developing enhanced versions through integration with NovelD [29] and DEIR [24] – state-of-the-art (SOTA) algorithms for MiniGrid – with experimental validation in MiniGrid-MA confirming MIR's superiority over baseline SOTA methods.
- Comprehensive empirical evidence showing performance gains through mutual reward integration, particularly in cooperative tasks requiring coordinated exploration strategies.

## 2    Related Work

This section reviews related work in two main parts. The first part discusses existing intrinsic reward methods. These methods can be categorized into prediction-based methods and exploration novelty-based methods. While single-agent exploration encompasses both categories, multi-agent exploration primarily relies on exploration reward strategies. These works will be discussed further, with some algorithms implemented in the experiments. The second part introduces a framework for constructing sparse discrete environments called MiniGrid. To evaluate exploration methods suitable for MARL, we will adapt this environment into a MARL variant termed MiniGrid-MA.

### 2.1    Exploration Methods

In single-agent reinforcement learning, extensive research efforts have been dedicated to intrinsic reward mechanisms for sparse-reward environments, demonstrating remarkable efficacy in exploration-driven tasks. Existing methods generally fall into two technical categories. The first approach integrates auxiliary models that process observational data from environment interactions to generate guiding intrinsic rewards, as exemplified by ICM [16] and NGU [2].

The second paradigm quantifies exploration novelty through state-visitation differentials, driving agents towards novel scenarios. Foundational work in this domain includes count-based methods [15] that quantify state novelty through empirical visitation counts. Later advancements extended this concept through unsupervised learning frameworks that process observations to develop auxiliary prediction tasks, ultimately generating compact state embeddings. These methods typically employ specialized encoders (e.g., CNN architectures or RNN modules) to achieve generalized environmental change detection. Representative implementations include RND [4], which measures discrepancy between fixed and trainable network outputs, and NovelD [29] which adapts RND for observation-space analysis. Most recently, DEIR [24] synergizes predictive modeling with environmental dynamics to achieve state-of-the-art performance on MiniGrid sparse-reward benchmarks.

Notably, recent advancements in MARL with sparse rewards have produced novel algorithms including I-Go-Explore [10], AgentTime [20], and lazyAgent [12]. The field also observes design innovations in multi-agent reward engineering, such as IRAT's [26] individual-team reward balancing framework. While our research design draws inspiration from these foundational works, this study specifically investigates how intrinsic reward mechanisms – particularly our Mutual Intrinsic Reward (MIR) paradigm can enhance existing MARL methods' performance in sparse-reward scenarios.

### 2.2    MiniGrid Environments

The MiniGrid [5] framework features customizable 2D gridworld environments where discrete cells represent interactive objects including walls, keys, tiles, and doors. Agents execute discrete actions comprising directional rotation, forward movement, object manipulation (pickup/drop), and door operation. A key characteristic is the environment's strictly deterministic state transitions, ensuring identical state-action pairs yield consistent outcomes. These maze environments naturally embody sparse-reward dynamics while maintaining intuitive visual representations, making them ideally suited for constructing controlled experimental benchmarks. MiniGrid has been extensively benchmarked for evaluating intrinsic reward algorithms such as NGU [2], NovelD [29], and DEIR [24].

To address the current limitation of lacking native multi-agent support, this work develops MiniGrid-MA – a multi-agent extension that introduces new cell object types and specialized grid configurations for cooperative task evaluation. This adaptation enhances evaluation capabilities for sparse-reward MARL scenarios through three key modifications: (1) Multi-agent action space definition, (2) Shared object interaction protocols, and (3) Collaborative reward attribution mechanisms.

## 3    Proposed Method and Environment

### 3.1    Framework in Multi-Agent Environment

Figure 2 presents our framework architecture. The black components represent the baseline MAPPO implementation under the CTDE paradigm, streamlined by omitting standard components (e.g., return/advantage calculations and loss term derivations) for visual clarity. Blue elements depict conventional intrinsic reward operations, as operationalized in DEIR [24], which incorporates auxiliary prediction tasks like temporal sequence classification of agent observations.

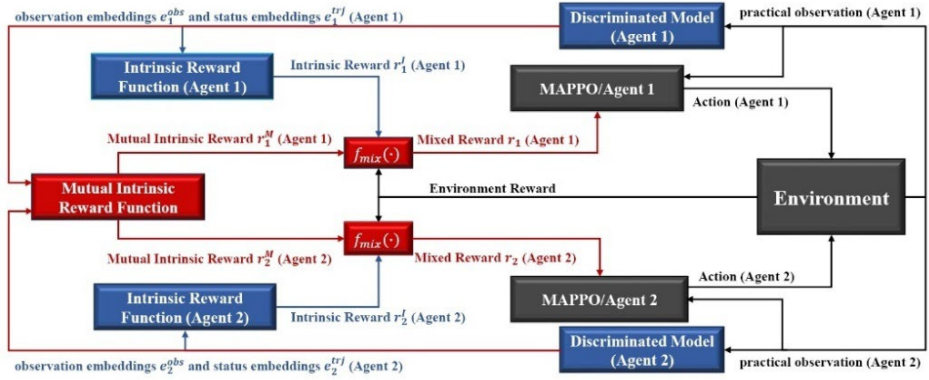

**Fig. 2.** Framework of the proposed method (in two agents' case)

For intrinsic reward computation, we exclusively utilize the discriminator's encoder module (CNN-GRU architecture). While our discrimination model originates from DEIR's design, we implement agent-specific discriminators with independent parameter updates.

Red components highlight our novel mutual reward mechanism enhancing inter-agent collaborative exploration. Crucially, our framework maintains compatibility with established intrinsic reward methods including ICM [16], NGU [2], NovelD [29], and DEIR [24], though DEIR and NovelD demonstrate higher architectural compatibility. The core innovation involves exploiting inter-agent influence patterns through reward coupling, which requires supplementary intrinsic reward modeling (Figure 3) to learn environment-agnostic observation embeddings.

### 3.2 Mutual Intrinsic Reward

In all methods utilizing intrinsic rewards mentioned in this paper, the reward for each agent consists of two components. The first component is the environmental reward, which drives the agent or team to achieve specified goals. The second component encompasses supplement rewards generated by different policies (termed supplement rewards here to differentiate from intrinsic rewards in single-agent settings), which assist the agent or team in a thorough exploration of the environment.

$$r_{(k,t)} = k_E * r_{(t)}^E + k_S * r_{(k,t)}^S, \tag{1}$$

where $r_{(k,t)}$ denotes the reward for agent $k$ at time $t$, $r_{(t)}^E$ is the environmental reward, $k_E$ is the environmental reward coefficient, $k_S$ is the supplement reward coefficient, and $r_{(k,t)}^S$ represents the supplement reward.

The supplement reward strategy designed for experimentation includes two key components. The first component employs the intrinsic reward from the DEIR for single-agent environments. The second component is the proposed mutual intrinsic reward (MIR) after a mixing function is applied.

$$r_{(k,t)}^S = k_I * r_{(k,t)}^I + k_M * f_{mix}\big(r_{(k,t)}^M\big), \tag{2}$$

where $r_{(k,t)}^S$ is the supplement reward for agent $k$ at time $t$, $r_{(k,t)}^I$ is the intrinsic reward, $k_I$ is the intrinsic reward coefficient, $r_{(k,t)}^M$ is the mutual intrinsic reward, $k_M$ is the mutual intrinsic reward coefficient, and $f_{mix}(\cdots)$ is the mix function.

In the calculation methods of intrinsic rewards, we consider two state-of-the-art (SOTA) methods: the DEIR method and the NovelD method. The core idea of the DEIR method is to train an additional discrimination model to learn whether any two frames are consecutive, thereby obtaining an encoder for the agent's observations, which is used to measure the novelty of the agent's observations. On the other hand, the core idea of NovelD is to directly use a Random Distillation Network (RND) to obtain the novelty of the agent's observations. Both methods employ similar CNN-based networks and GRU networks to extract spatial and temporal features.

In our method, we utilize both the DEIR method [24] and the NovelD method [29] to generate intrinsic rewards, and then, based on the principles of MIR, we design the MIR computation formula. The intrinsic reward from DEIR method [24] for our environments can be expressed as follows.

$$r_{(k,t)}^I = \min_{\forall i \in [0,t]} \frac{\text{dist}^2\left(e_{(k,i)}^{obs}, e_{(k,t+1)}^{obs}\right)}{\text{dist}\left(e_{(k,i)}^{trj}, e_{(k,t)}^{trj}\right) + \epsilon_m}, \tag{3}$$

where dist $(\cdot)$ is the distance function, using Euclidean distance to measure the distance between different embedded representations. $r_{(k,t)}^I$ is the intrinsic reward for agent $k$ at time $t$ and $\epsilon_m$ is a small constant to address numerical stability. $e_{(k,i)}^{obs}$ represents the observation encoding for agent $k$ at time $i$ (extracted through the CNN module), and $e_{(k,i)}^{trj}$ denotes the trajectory encoding (extracted through the GRU module). These embeddings are obtained from encoder modules (CNN and GRU) in the discriminated model (Figure 3), derived from DEIR. The loss function of the discrimination model is the prediction error of positive and negative sample pairs.

The intrinsic reward from NovelD method [29] for our environments can be expressed as follows.

$$r_{(k,t)}^I = \max\left[\text{novelty}\left(e_{(k,t)}^{trj}\right) - \alpha * \text{novelty}\left(e_{(k,t-1)}^{trj}\right), 0\right], \tag{4}$$

where novelty$(s\_t) = ||\phi_{\text{fixed}}(e_{(k,t)}^{trj}) - \phi_{\text{train}}(e_{(k,t)}^{trj})||$, To measure the novelty of the state in large-scale stochastic environments Similar to the RND method [4].

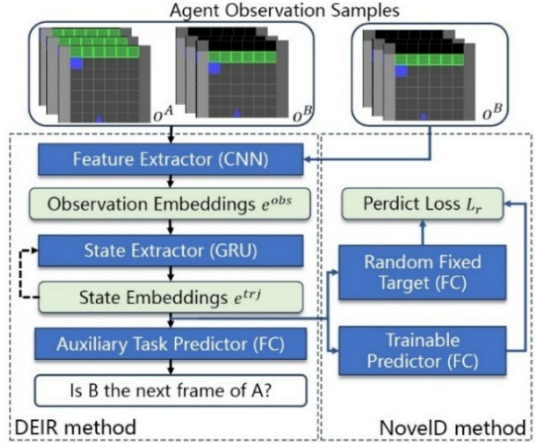

**Fig. 3.** The Structure of the Discriminated Model

When designing mutual intrinsic rewards through intrinsic rewards, the motivation is relatively straightforward. To evaluate the influence of individual agents on their teammates, we introduce rewards that encourage exploration behaviors capable of altering the observations of other agents within the team.

For the DEIR-based calculation of mutual intrinsic rewards, we directly utilize the encoder learned from the DEIR method to encode agent observations. The mutual reward is computed by assessing the novelty of other agents' observation encodings within their historical sequences. Specifically, the DEIR-based mutual intrinsic reward [24] is calculated using the encoded observations of other team members.

$$r_{(k,t)}^M = \max_{\forall j \in [0,K], j \neq k} \min_{\forall i \in [0,t)} \text{dist}^2\left(e_{(j,i)}^{obs}, e_{(j,t+1)}^{obs}\right) \tag{5}$$

where $r_{(k,t)}^M$ is the mutual intrinsic reward for agent $k$ at time $t$.

When calculating mutual intrinsic rewards using the NovelD method, we directly utilize the Random Distillation Network's prediction error from NovelD to determine observation novelty of other agents. The mutual reward is obtained by computing prediction errors from teammates' Random Distillation Networks. Specifically, the NovelD-based mutual intrinsic reward [29] is determined through encoded representations of teammates' observations.

$$r_{(k,t)}^M = \max\left[\max_{\forall j \in [0,K], j \neq k} \text{novelty}\left(e_{(j,t)}^{trj}\right) - \alpha * \text{novelty}\left(e_{(j,t-1)}^{trj}\right), 0\right], \tag{6}$$

here, $r_{(k,t)}^M$ represents the mutual intrinsic reward for agent $k$ at time $t$. The novelty of state $s_t$ is defined as:

$$\text{novelty}(s_t) = \left\lVert \phi_{\text{fixed}}\left(e_{(k,t)}^{trj}\right) - \phi_{\text{train}}\left(e_{(k,t)}^{trj}\right)\right\rVert, \tag{7}$$

where $\phi_{\text{fixed}}$ denotes a fixed-parameter, non-trainable random encoding network, and $\phi_{\text{train}}$ represents a trainable predictive network. As environmental exploration and training progress, $\phi_{\text{train}}$ becomes better at predicting the encodings generated by $\phi_{\text{fixed}}$ for observed states or trajectories. The prediction error between these two networks serves as a metric to quantify the novelty of observations or states.

After determining the intrinsic and mutual intrinsic rewards, we apply the intrinsic reward as a weight using the softmax function to obtain the reweighted mutual intrinsic reward. This ensures that only those agents with relatively high intrinsic rewards within the team can achieve high mutual intrinsic rewards, preventing all agents from being driven solely by mutual intrinsic rewards and neglecting individual exploration. The specific computation is given as follows.

$$f_{mix}\left(r_{(k,t)}^M\right) = \frac{\exp\left(\left(r_{(k,t)}^I - \max_{i \in [1,K]} r_{(i,t)}^I\right)/T\right)}{\sum_{k=1}^K \exp\left(\left(r_{(k,t)}^I - \max_{i \in [1,K]} r_{(i,t)}^I\right)/T\right)} * r_{(k,t)}^M, \tag{8}$$

where $f_{mix}\left(r_{(k,t)}^M\right)$ represents the weighted mutual intrinsic reward for agent $k$ at time $t$. To maintain stability in rewards, we employ a similar momentum normalization strategy as used in DEIR [24] and Value Norm [28].

Based on the aforementioned framework and the methods for calculating different rewards, we outline the complete pseudo-code (Algorithm 1) for the proposed algorithm framework. In the pseudocode, we provide a detailed specification of the MAPPO algorithm components relevant to our MIR implementation, along with the precise computational workflow of MIR. We also include a concise overview of the MAPPO model's gradient computation and update procedures, which remain consistent with the original algorithm. Notably, our proposed method is not a fixed reward strategy. MIR can be flexibly integrated with various existing intrinsic reward methods. This integration aims to enhance the performance of intrinsic reward methods originally designed for single-agent reinforcement learning tasks when applied to multi-agent reinforcement learning environments. Furthermore, task-specific weighting coefficients or hybrid strategies can be incorporated to balance individual exploration incentives and team coordination requirements in cooperative scenarios.

### 3.3    MiniGrid-MA Environments

MiniGrid [5] provides customizable maze environments, and many intrinsic reward methods and training algorithms designed for sparse reward scenarios have utilized this environment for testing. However, the current version only supports single-agent functionality, necessitating an extension we refer to as MiniGrid-MA. We acknowledge that we are not the first team to propose this idea.

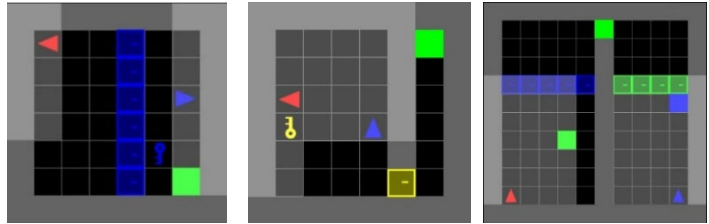

**Fig. 4.** Minigrid-MA environment (DoorKeyB, DoorSwitchA and DoorSwitchB)

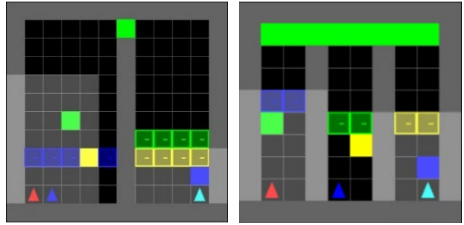

**Fig. 5.** Minigrid-MA environment (DoorSwitchC and DoorSwitchD)

**Table 1.** Information on the Maps of MiniGrid-MA

| Type | Teamwork | Agents | Map Size |
|------|----------|--------|----------|
| DoorKeyB | YES | 2 | 6 and 8 |
| DoorSwtichA | YES | 2 | 8, 12 and 16 |
| DoorSwtichB | YES | 2 | 8, 10 and 16 |
| DoorSwtichC | YES | 3 | 8, 12 and 16 |
| DoorSwtichD | YES | 3 | 10, 12 and 14 |

Previous research [20] has introduced multiple agents into MiniGrid, experimenting with two maps, one of which, DoorKeyA (with size 8), is identical to that in earlier work. Building upon this foundation, we have designed additional multi-agent reinforcement learning (MARL) environments that require teamwork and involve sparse rewards to evaluate the performance of our method alongside others. We have also varied map sizes and the number of agents to obtain more reliable experimental results. A summary of these maps is presented in Table 1.

In all environments, a team reward related to completion time is awarded only when all agents reach the target point; no other external rewards are given at any other time. The specific reward can be calculated using the formula as follows.

$$r(t) = \begin{cases} 2 - \frac{t}{T}, & \text{If task is completed} \\ 0, & \text{Others} \end{cases}, \tag{9}$$

here, $t$ represents the time taken for all agents to reach the target, and $T$ is the predefined maximum time. By incorporating this time-dependent factor, we encourage the agent team to complete tasks as quickly as possible.

## 4  Experiments

It is important to note that the environments used in this study only provide external rewards when all agents fully complete the team task. The external reward is not offered during other phases. Our experiments primarily focus on
the following questions, each accompanied by corresponding experiments.

1. How does our method (DEIR with MIR) perform compared to others in sparse reward multi-agent environments that require teamwork to complete tasks?
2. How does our method (NovelD with MIR) perform compared to others in sparse reward multi-agent environments that require teamwork to complete tasks?

Additionally, we conducted several ablation experiments to assess whether the specific strategies we proposed effectively address these issues, divided into two main sections: the mixing method for intrinsic and mutual intrinsic rewards and the MAPPO techniques used in this study.

**Parameters Settings** To ensure experimental comparability, we maintained consistent algorithmic parameter settings across different maps. Due to space limitations. Specifically, the network architectures employed in our algorithm, including the discriminator from DEIR [24] and the random fixed target model with trainable predictor from NovelD [29]. The parameter settings for the MAPPO algorithm, intrinsic reward strategy, and mutual intrinsic reward strategy used in our experiments are available in the "Parameter Settings" section. During experiments, we assigned different random seed sequences for varying numbers of runs and sampling threads to generate diverse map configurations. This approach ensures the diversity and robustness of experimental results while preventing potential biases from specific map configurations. Through this methodology, we achieve comprehensive evaluation of algorithm performance across different environments.

**Experimental Result** In this experiment, we aimed to evaluate the performance of our method compared to other approaches in sparse reward multi-agent environments that necessitate teamwork to accomplish tasks. Accordingly, we focused on designing maps that require collaboration for task completion. We considered maps such as DoorKeyB, DoorKeyC, DoorSwitchA, DoorSwitchB, DoorSwitchC and DoorSwitchD with varying sizes to assess the performance of various intrinsic reward algorithms. The results for the NovelD method, NovelD with MIR method, DEIR method, DEIR algorithm with MIR method , and other methods are illustrated in Table 2. Each cell represents the highest average reward obtained by the method across different runs under the corresponding map. Bold text indicates the best performance among all algorithms, and an upward arrow signifies that the method with MIR outperforms the original method without the MIR strategy.

**Table 2.** Mean Best Episode Reward in Experiment (NovelD-MIR part)

| Environment | MAPPO (No-Model) | RND | NGU | NovelD | NovelD-MIR (Ours) | DEIR | DEIR-MIR (Ours) |
|---|---|---|---|---|---|---|---|
| DoorKeyB6x6 | 0.80 | 0.24 | 0.25 | 1.32 | **1.97**↑ | 1.61 | 1.75↑ |
| DoorKeyB8x8 | 0.05 | 0.04 | 0.08 | 0.91 | **1.15**↑ | 0.11 | 0.39↑ |
| DoorSwitch8x8 | 0.59 | 0.22 | 0.40 | 1.42 | 1.33 | 0.40 | **1.54**↑ |
| DoorSwitchB8x8 | 0.56 | 0.18 | 0.14 | **1.27** | 1.22 | 0.86 | 0.91↑ |
| DoorSwitchC8x8 | 0.03 | 0.03 | 0.05 | **0.86** | 0.60 | 0.08 | 0.36↑ |
| DoorSwitchB10x10 | 0.06 | 0.01 | 0.05 | **0.85** | 0.77 | 0.16 | 0.62↑ |
| DoorSwitchD10x10 | 0.00 | 0.03 | 0.01 | 0.25 | **0.45**↑ | 0.07 | 0.25↑ |
| DoorSwitch12x12 | 0.01 | 0.01 | 0.08 | 0.94 | 1.26↑ | 0.13 | **1.44**↑ |
| DoorSwitchB12x12 | 0.00 | 0.01 | 0.00 | 0.13 | **0.58**↑ | 0.08 | 0.48↑ |
| DoorSwitchC12x12 | 0.00 | 0.00 | 0.00 | **0.03** | 0.03 | 0.01 | **0.04**↑ |
| DoorSwitchD12x12 | 0.01 | 0.00 | 0.00 | 0.05 | **0.07**↑ | 0.00 | 0.01↑ |
| DoorSwitchD14x14 | 0.00 | 0.00 | 0.00 | 0.02 | **0.11**↑ | 0.01 | 0.00 |
| DoorSwitch16x16 | 0.04 | 0.00 | 0.13 | 0.44 | **1.25**↑ | 0.07 | 0.48↑ |
| DoorSwitchB16x16 | 0.01 | 0.00 | 0.00 | 0.03 | **0.42**↑ | 0.00 | 0.18↑ |
| DoorSwitchC16x16 | 0.00 | 0.00 | 0.00 | 0.00 | 0.01↑ | 0.00 | **0.03**↑ |

Based on the experimental results presented above, we can draw several straightforward conclusions. Firstly, intrinsic reward SOTA methods that perform well in sparse-reward single-agent environments, such as the Discriminative-model-based DEIR method [24] or the Random Network Distillation-based NovelD method [29], can enhance their performance in sparse-reward multi-agent environments by incorporating the MIR strategy. This enhancement strategy does not require the introduction of additional models to be trained, compared to the original methods. Secondly, in multi-agent environments with sparse rewards, as the scale of the map increases or the difficulty level rises, merely employing the MAPPO algorithm or directly applying intrinsic reward methods like DEIR or NovelD to each agent does not enable the agent team to effectively explore and obtain rewards, thus hindering normal training progress.

Thus, we conclude that in sparse reward multi-agent environments requiring teamwork, our method is more adept at focusing on exploratory cooperative actions, thereby enhancing the exploration efficiency of the agent team.

# 5    Discussion & Future Work

Mutual Intrinsic reward (MIR) is not mutually exclusive to existing rewards. MIR can be integrated with various intrinsic reward methods to enhance policy performance in multi-agent environments. MIR can effectively improve the performance of intrinsic reward methods originally designed for single-agent reinforcement learning tasks when applied to multi-agent reinforcement learning tasks. Additionally, different weighting strategies can be introduced for various tasks to balance exploration in single-agent environments with joint exploration in collaborative scenarios, presenting avenues for further research.

However, MIR also has limitations. For certain hard tasks that do not require collaboration (even though most tasks do), this approach may compromise the performance of standard algorithms. Therefore, well-designed weighting strategies are needed to assess the relationship between intrinsic rewards and MIR. Furthermore, the current methodology is hard to directly reward joint actions that long-term modify the observations of other agents.

In future work, we plan to develop MIR strategies based on this concept and integrate them with various intrinsic reward designs, aiming to create more advanced strategies for different scenarios.

**Acknowledgments.** This study is supported by Shenzhen Fundamental Research Program (Grant No. JCYJ20220818102414030), the Major Key Project of PCL (Grant No. PCL2024A05), Shenzhen Science and Technology Program (Grant No. ZDSYS20210623091809029), Guangdong Provincial Key Laboratory of Novel Security Intelligence Technologies (Grant No. 2022B1212010005).